\documentclass[]{bytedance_seed}

\usepackage[toc,page,header]{appendix}

\usepackage{minitoc}
\usepackage{algorithm}
\usepackage{algpseudocode}
\usepackage{enumitem}
\usepackage{xcolor}
\usepackage{amsfonts}
\usepackage{amssymb}

\title{What's Behind PPO's Collapse in Long-CoT? \\ 
Value Optimization Holds the Secret}

\author[]{Yufeng Yuan}
\author[]{Yu Yue}
\author[]{Ruofei Zhu}
\author[]{Tiantian Fan}
\author[]{Lin Yan}

\affiliation[]{ByteDance Seed}

\abstract{

Reinforcement learning (RL) is pivotal for enabling large language models (LLMs) to generate long chains of thought (CoT) for complex tasks like math and reasoning. However, Proximal Policy Optimization (PPO), effective in many RL scenarios, fails in long CoT tasks. This paper identifies that value initialization bias and reward signal decay are the root causes of PPO's failure. We propose Value-Calibrated PPO (VC-PPO) to address these issues. In VC-PPO, the value model is pretrained to tackle initialization bias, and the Generalized Advantage Estimation (GAE) computation is decoupled between the actor and critic to mitigate reward signal decay. Experiments on the American Invitational Mathematics Examination (AIME) show that VC-PPO significantly boosts PPO performance. Ablation studies show that techniques in VC-PPO are essential in enhancing PPO for long CoT tasks.
}

\date{\today}
\correspondence{Yufeng Yuan at \email{yufeng.yuan@bytedance.com}}

\begin{document}
\maketitle


\section{Introduction}

In recent years, large language models (LLMs) have achieved remarkable breakthroughs across diverse domains, including question-answering \citep{han2024psydial}, code generation \citep{jimenez2023swe, chen2024step}, dialog generation \citep{han2024psydial}, and agent-related tasks \citep{wang2024opendevin}. A particularly notable advancement is the ability of LLMs to solve Olympiad-level math and reasoning problems. This feat is accomplished by generating a long Chain of Thought (CoT) \citep{cot_2022} before reaching a final answer. Such an inference-time scaling paradigm was initially proposed by OpenAI-o1 \citep{o1} and further popularized by DeepSeek-R1 \citep{deepseekai2025deepseekr1incentivizingreasoningcapability} and OpenAI-o3 \citep{o3}. During the long CoT process, the model formulates hypotheses and verifies them to gradually converge to a correct solution.

To equip models with such capabilities, researchers typically follow these procedures:

\begin{enumerate}[leftmargin=*]
    \item \textbf{Cold-start with Supervised Fine-tuning (SFT) data}: A substantial amount of SFT data following the Long-CoT pattern is collected. This initial step equips the model with a fundamental understanding of how to test and verify its own answers, laying a solid foundation for subsequent learning.
    \item \textbf{Construct a dataset with verifiable tasks}: The dataset is composed of tasks such as math and reasoning problems, whose correctness can be objectively judged by a non-hackable, rule-based reward model. This ensures that the model receives reliable feedback during training.
    \item \textbf{Reinforcement learning (RL) training}: The model is trained using RL with the objective of maximizing the reward from the rule-based reward model. Through this process, the model's long CoT capabilities are solidified and enhanced, leading to a significant improvement in its performance on complex tasks.
\end{enumerate}

Reinforcement learning has played a critical role in developing such capabilities. However, directly applying Proximal Policy Optimization (PPO) \citep{schulman2017proximal}, a method that has proven effective in various fields, including Reinforcement Learning with Human Feedback (RLHF), can lead to failure modes in tasks that require long CoT. As the response length increases, obtaining an accurate value model becomes increasingly challenging both before and during training. In contrast, Group Relative Policy Optimization (GRPO) \citep{shao2024deepseekmath}, a simplified version of PPO that replaces the value model with the Leave-One-Out \citep{rloo} estimate, has shown strong performance in such tasks. However, compared to GRPO, which only uses response-level feedback, PPO can utilize more fine-grained token-level feedback, indicating that PPO could have higher potential in complex tasks that require extensive exploration.

In this paper, our goal is to fully exploit the potential of PPO in long CoT tasks. We first identify the key problem of PPO in long CoT tasks: the value model exhibits considerable bias before and during training, which causes it to fail to predict values accurately. The pre-training value bias stems from the common practice of initializing the value model from the reward model. During the early training stage, this approach leads to a large error in advantage estimation. The in-training value bias arises from the decaying nature of Generalized Advantage Estimation (GAE) \citep{schulman2015high} computation. In scenarios with long sequences and rewards at the end, the value function fails to propagate the reward signal to the preceding tokens.

To address the value bias in PPO, we propose Value-Calibrated PPO (VC-PPO), in which the value model is calibrated before and during training. To address the value initialization bias, we pretrain the value model with responses generated by a fixed SFT policy in an offline manner. This helps the value model to better estimate the expected rewards and reduces the bias in the early training phase. To mitigate the value bias during training, we propose to decouple the GAE computation for the policy and the value, such that the value could use a larger $\lambda$ to allow a more effective propagation of the reward signal along the long sequence, while the policy could maintain the original $\lambda$ to ensure convergence under time and computational constraints.

In our experiments on the American Invitational Mathematics Examination (AIME), these two techniques significantly boost the performance of the baseline PPO from 5.6 to 49.0, achieving a higher score than previously reported in \citep{deepseekai2025deepseekr1incentivizingreasoningcapability}. Moreover, our ablation studies demonstrate that both techniques are essential for achieving superior performance in AIME, highlighting the importance of our proposed solutions in enhancing the effectiveness of PPO in Long-CoT tasks.

\section{Preliminaries}
This section presents the fundamental concepts and notations that serve as the basis for our proposed algorithm. We first explore the basic framework of representing language generation as a reinforcement learning task. Subsequently, we introduce Proximal Policy Optimization and Generalized Advantage Estimation.

\subsection{Modeling Language Generation as Token-Level MDP}
Reinforcement Learning (RL) centers around the learning of a policy that maximizes the cumulative reward for an agent as it interacts with an environment. In this study, we cast language generation tasks within the framework of a Markov Decision Process (MDP) \citep{ouyang2022training}.

Let the prompt be denoted as $x$, and the response to this prompt as $y$. Both $x$ and $y$ can be decomposed into sequences of tokens. For example, the prompt $x$ can be expressed as $x=(x_0,\dots,x_m)$, where the tokens are drawn from a fixed discrete vocabulary $\mathcal{A}$.

We define the token-level MDP as the tuple $\mathcal{M}=(\mathcal{S},\mathcal{A},\mathbb{P},r,d_0,\omega)$. Here is a detailed breakdown of each component:
\begin{itemize}[leftmargin=*]
    \item \textbf{State Space ($\mathcal{S}$)}: This space encompasses all possible states formed by the tokens generated up to a given time step. At time step $t$, the state $s_t$ is defined as $s_t=(x_0,\dots,x_m,y_0,\dots,y_t)$.
    \item \textbf{Action Space ($\mathcal{A}$)}: It corresponds to the fixed discrete vocabulary, from which tokens are selected during the generation process.
    \item \textbf{Dynamics ($\mathbb{P}$)}: These represent a deterministic transition model between tokens. Given a state $s_t=(x_0,\dots,x_m,y_0,\dots,y_t)$, an action $a = y_{t + 1}$, and the subsequent state $s_{t+1}=(x_0,\dots,x_m,y_0,\dots,y_t,y_{t+1})$, the probability $\mathbb{P}(s_{t+1}|s_t,a)=1$.
    \item \textbf{Termination Condition}: The language generation process concludes when the terminal action $\omega$, typically the end-of-sentence token, is executed.
    \item \textbf{Reward Function ($r(s,a)$)}: This function offers scalar feedback to evaluate the agent's performance after taking action $a$ in state $s$. In the context of Reinforcement Learning from Human Feedback (RLHF), the reward function can be learned from human preferences or defined by a set of rules specific to the task.
    \item \textbf{Initial State Distribution ($d_0$)}: It is a probability distribution over prompts $x$. An initial state $s_0$ consists of the tokens within the prompt $x$.
\end{itemize}

\subsection{RLHF Learning Objective}
We formulate the optimization problem as a KL-regularized RL task. Our objective is to approximate the optimal KL-regularized policy, which is given by:
\begin{align}\label{eq:objective}
   \pi^* =  \arg\max_\pi \mathbb{E}_{\pi, s_0 \sim d_0} \left[ \sum_{h = 0}^H  \left(r(s_h, a_h)-\beta \text{KL} \big( \pi(\cdot | s_h) \| \pi_{\text{ref}}(\cdot | s_h) \big)\right) \right]
\end{align}
In this equation, $H$ represents the total number of decision steps, $s_0$ is a prompt sampled from the dataset, $r(s_h, a_h)$ is the token-level reward obtained from the reward function, $\beta$ is a coefficient that controls the strength of the KL-regularization, and $\pi_{\text{ref}}$ is the initialization policy.

In traditional RLHF and most tasks related to Large Language Models (LLMs), the reward is sparse and is only assigned at the terminal action $\omega$, that is, the end-of-sentence token \texttt{<eos>}. 

\subsection{Proximal Policy Optimization}
PPO \citep{schulman2017proximal} uses a clipped surrogate objective to update the policy. The key idea is to limit the change in the policy during each update step, preventing large policy updates that could lead to instability.

Let $\pi_{\theta}(a|s)$ be the policy parameterized by $\theta$, and $\pi_{\theta_{\text{old}}}(a|s)$ be the old policy from the previous iteration. The surrogate objective function for PPO is defined as:

\begin{equation}
\mathcal{L}^{CLIP}(\theta)=\hat{\mathbb{E}}_t\left[\min\left(r_t(\theta)\hat{A}_t,\text{clip}(r_t(\theta), 1-\epsilon, 1+\epsilon)\hat{A}_t\right)\right]
\end{equation}

where $r_t(\theta)=\frac{\pi_{\theta}(a_t|s_t)}{\pi_{\theta_{\text{old}}}(a_t|s_t)}$ is the probability ratio, $\hat{A}_t$ is the estimated advantage at time step $t$, and $\epsilon$ is a hyperparameter that controls the clipping range.

Generalized Advantage Estimation (GAE) \citep{schulman2015high} is a technique used to estimate the advantage function more accurately in PPO. It combines multiple-step bootstrapping to reduce the variance of the advantage estimates. For a trajectory of length $T$, the advantage estimate $\hat{A}_t$ at time step $t$ is computed as:

\begin{equation}
\hat{A}_t=\sum_{l = 0}^{T-t-1}(\gamma\lambda)^l\delta_{t + l}
\label{eq:gae_definition}
\end{equation}

where $\gamma$ is the discount factor, $\lambda\in[0, 1]$ is the GAE parameter, and $\delta_t=r_t+\gamma V(s_{t + 1})-V(s_t)$ is the temporal-difference (TD) error. Here, $r_t$ is the reward at time step $t$, and $V(s)$ is the value function. Since it is a common practice to use discount factor $\gamma = 1.0$ in RLHF, to simplify our notation, we omit $\gamma$ in later sections of this paper.
\section{Identifying and Addressing PPO's Failure Modes in Long CoT Tasks}

In this section, we show a common failure mode of PPO in long CoT tasks and examine its relationship with the pre-training and in-training value biases from both theoretical and empirical perspectives. Subsequently, we propose practical solutions to enhance PPO and enable it to avoid such failures.

\subsection{Failure Modes of PPO in Long CoT Tasks}

Two common practices when applying PPO in the domain of Reinforcement Learning from Human Feedback (RLHF) are as follows \citep{zheng2023secretsrlhflargelanguage, huang2024nimplementationdetailsrlhf}:
\begin{itemize}[leftmargin=*]
    \item Employ the default Generalized Advantage Estimation (GAE), typically with $\lambda = 0.95$.
    \item Initialize the value model using a well-trained reward model.
\end{itemize}

\begin{figure}[htbp]
    \centering
    \begin{subfigure}[b]{0.45\textwidth}
        \centering
        \includegraphics[width=1.0\linewidth]{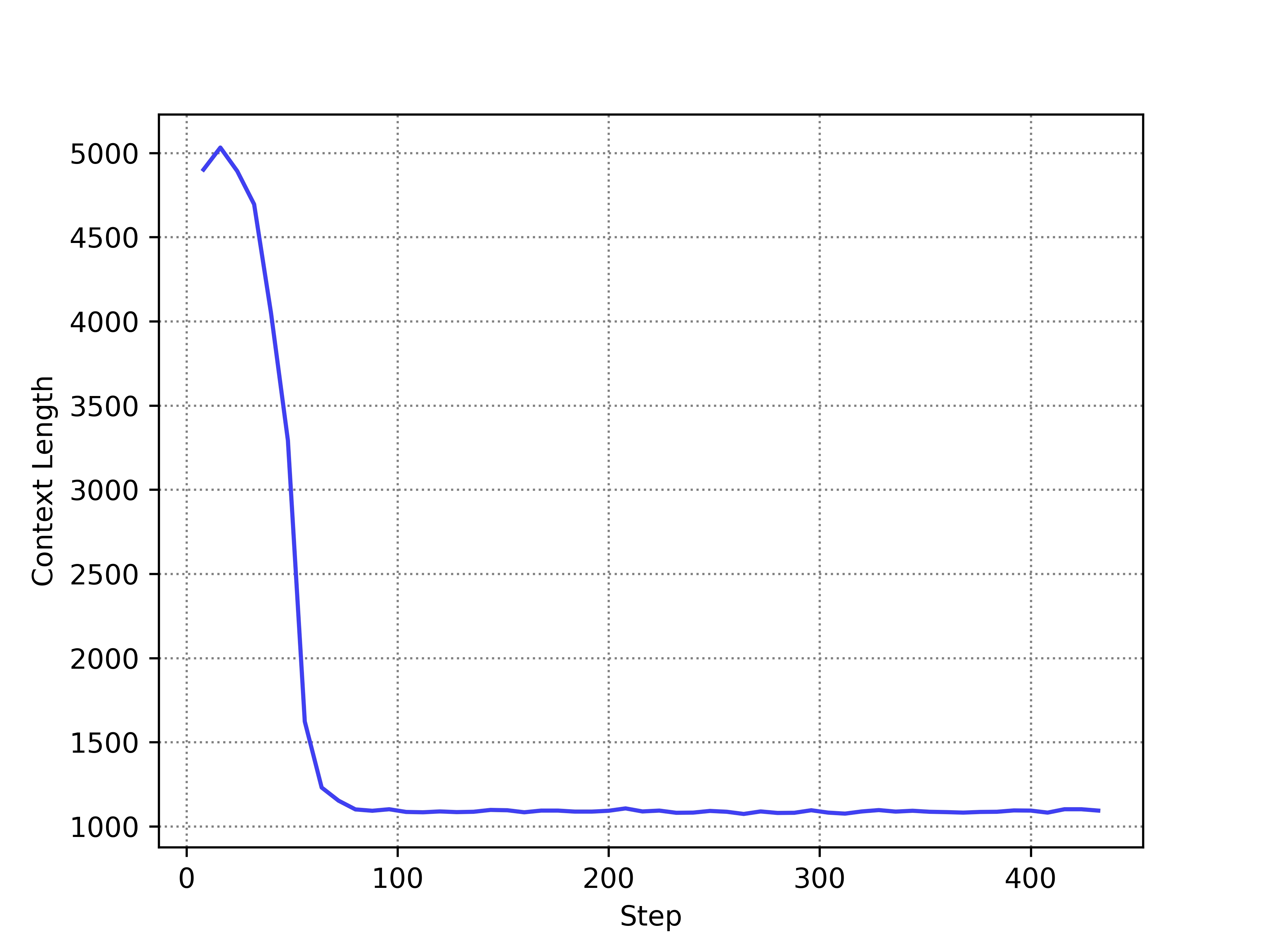}
        \caption{Model output length}
        \label{fig:ppo_failure_modes_1}
    \end{subfigure}
    \hfill
    \begin{subfigure}[b]{0.45\textwidth}
        \centering
        \includegraphics[width=1.0\linewidth]{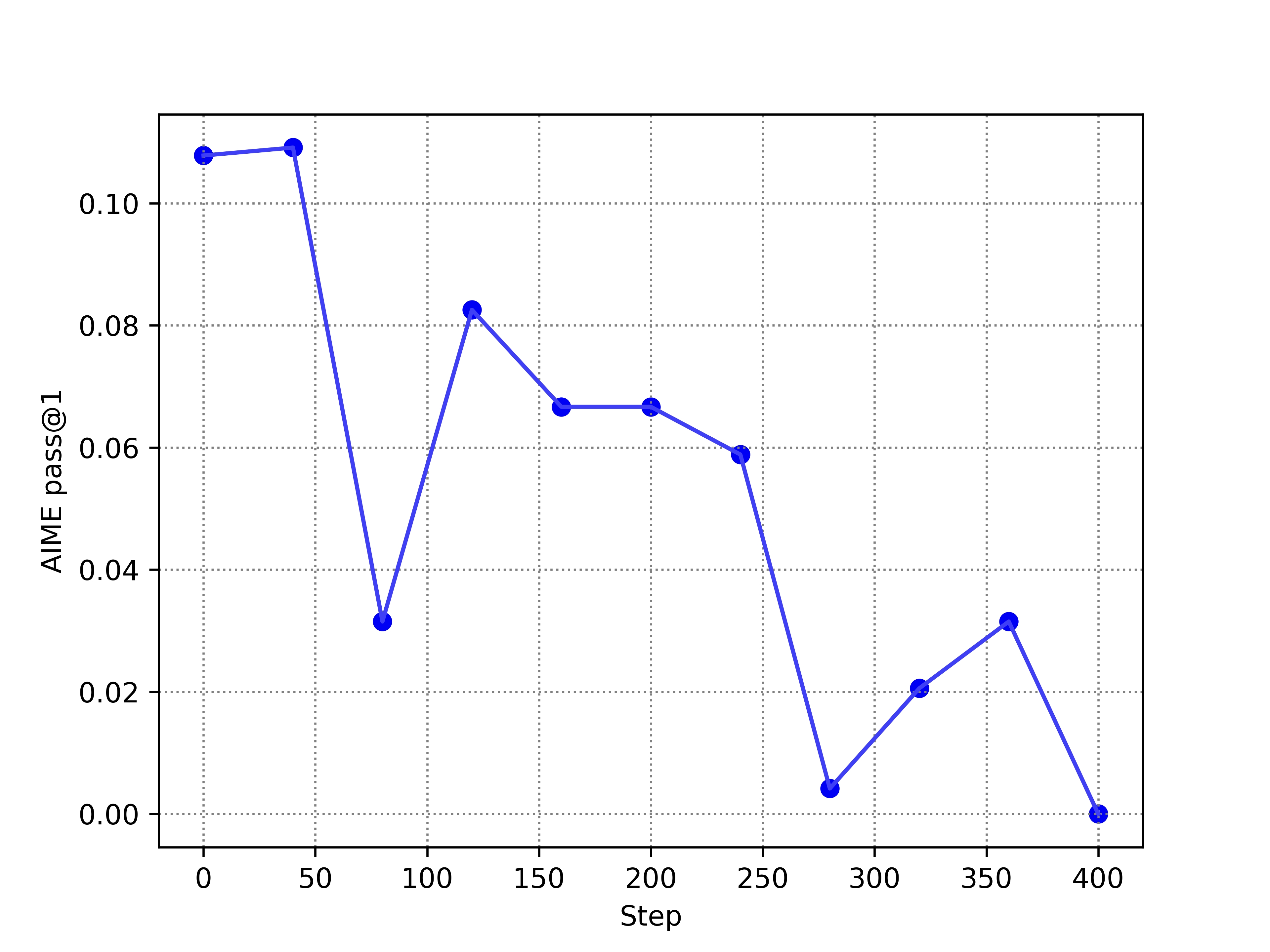}
        \caption{Validation performance on AIME}
        \label{fig:ppo_failure_modes_2}
    \end{subfigure}
    \caption{The failure modes of PPO observed in our experiments.}
    \label{fig:ppo_failure_modes}
\end{figure}

The first practice finds its origin in the traditional RL literature, where PPO has been extensively tested in environments like Mujoco \citep{conf/iros/TodorovET12} and Atari \citep{bellemare13arcade}. In these environments, the rewards accumulate over the trajectory, resulting in high-variance return. As a consequence, variance reduction becomes a necessity. The second practice emerges naturally from the apparent similarity between a reward model and a value model, since both models are trained to predict scalar information about the response. However, our experiments have revealed that naively applying PPO to tasks that require long CoT inevitably leads to failure, as shown in Figure \ref{fig:ppo_failure_modes}.

Typically, the failure modes are that the validation performance degrades as the training starts, accompanied by a significant decrease in the model's output length. Since it has been demonstrated that output length is strongly correlated with the model's performance on complex reasoning tasks \citep{o1}, the initial collapse in output length can be attributed as the root cause of this performance degradation.

\subsection{Addressing the Value Initialization Bias by Value Pretraining}

\begin{figure}[htbp]
    \centering
    \begin{subfigure}[b]{0.45\textwidth}
        \centering
        \begin{center}
            \includegraphics[width=1.0\linewidth]{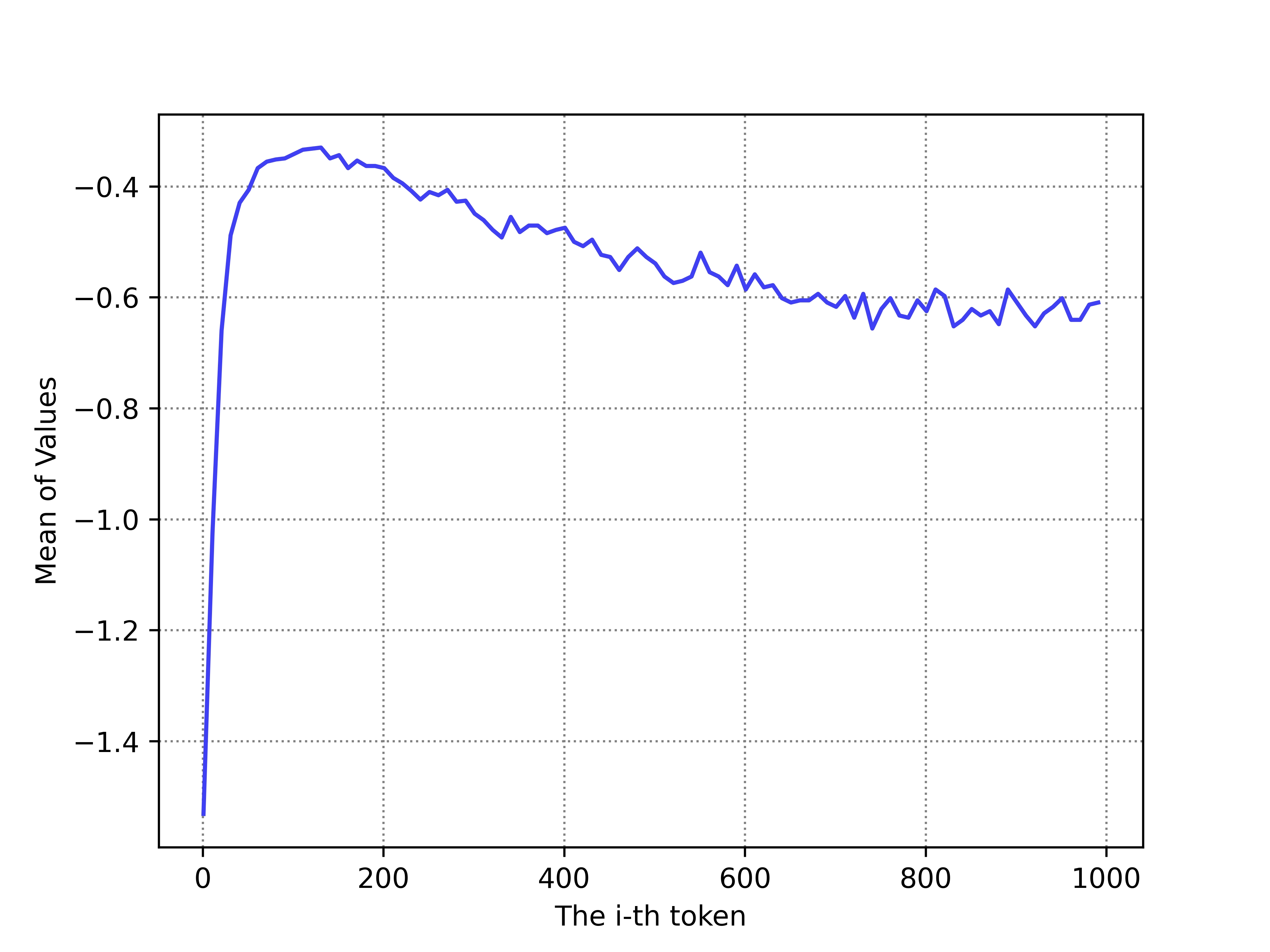}
        \end{center}
        \caption{Values at different token positions}
        \label{fig:ppo_failure_explained_1}
    \end{subfigure}
    \hfill
    \begin{subfigure}[b]{0.45\textwidth}
        \centering
        \begin{center}
            \includegraphics[width=1.0\linewidth]{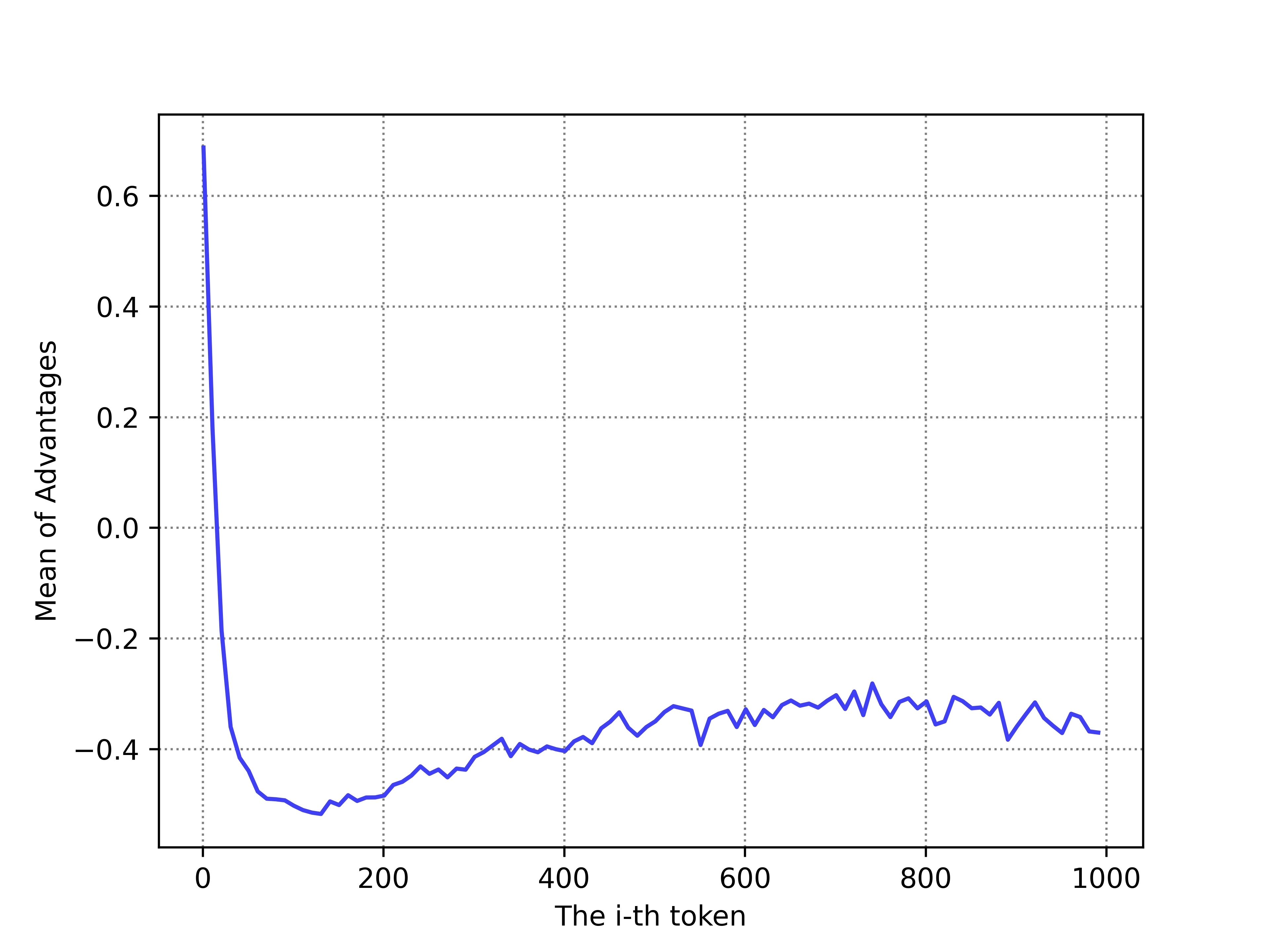}
        \end{center}
        \caption{Advantages at different token positions}
        \label{fig:ppo_failure_explained_2}
    \end{subfigure}
    \caption{Value and advantage bias with respect to token positions.}
    \label{fig:ppo_failure_explained}
\end{figure}

In our tasks, a verifier serves as the source of the reward signal. It utilizes a rule-based answer parsing mechanism, which is unlikely to show a preference for output length. Consequently, the reduction in output length can only be ascribed to the policy optimization dynamics, which are mainly driven by the advantages assigned to each token. To further explore this, we plot the correlation between advantages and the position of tokens, as shown in Figure \ref{fig:ppo_failure_explained}. This reveals a strong correlation between advantages and token position. The more preceding the tokens are, the more positively biased their advantages are. This causes the model to favor preceding tokens, ultimately leading to the observed collapse in output length.

The root cause of the positive bias lies in the objective mismatch between the reward model and the value model. The training objective of the reward model is to score the response at the \texttt{<EOS>} token. Since the tokens preceding \texttt{<EOS>} are not included in the training, the reward model tends to assign lower scores to tokens that are more preceding due to its increasing incompleteness. On the other hand, the aim of value prediction is to estimate the expected rewards of each token preceding \texttt{<EOS>} for a given policy. Given that tokens which are more preceding have lower scores and the KL-penalties are essentially zero at the beginning of training, there will be a positive bias at every timestep $t$ that accumulates along the trajectory, which is obvious through how advantages $\hat{A}_t$ is computed:

\begin{equation}
\hat{A}_t=\sum_{l = 0}^{T-t-1}\lambda^l (r_{t + l}+  V(s_{t + l + 1})-V(s_{t + l}))
\label{eq:gae_computation}
\end{equation}

This explains why tokens that are preceding tend to exhibit a greater positive bias in advantages. Due to this correlation between token position and advantages, the model tends to generate shorter responses, which prevents the model from generating a long chain of thoughts before finalizing an answer.

To alleviate such value initialization bias, we introduce \textbf{Value-Pretraining}. This approach involves offline training the value model until convergence under a pre-specified fixed policy. Once the value model has converged, it will be employed in all subsequent formal experiments. The specific steps are outlined as follows:

\begin{enumerate}[leftmargin=*]
    \item Continuously generate responses by sampling from a fixed policy, for instance, $\pi_{\text{sft}}$, and update the value model using GAE with $\lambda = 1.0$, also known as the Monte-Carlo return. This setting transforms the optimization problem into a stable gradient-descent optimization, ensuring more reliable and consistent updates to the value model.
    \item Train the value model until key training metrics, including value loss and explained variance \citep{explained_variance}, attain sufficiently low values. Monitoring these metrics is crucial as they reflect the quality and stability of the model's learning process, and reaching low values indicates that the model is converging effectively.
    \item Save the value checkpoint upon the completion of training. Subsequently, load this checkpoint for following experiments. This step provides a more accurate initial point for value estimation, enabling the model to start from a well-calibrated state.
\end{enumerate}

\subsection{Improving In-training Value Estimate with Decoupled-GAE}
\label{subsec:decoupled-gae}

Variance reduction is a critical topic in RL. The use of GAE with $\lambda = 0.95$ is common in traditional RL tasks like Mujoco and Atari, where accumulated rewards have high variance and lead to slow convergence. In contrast, in RLHF, a reward model or rule-based scoring mechanism offers trajectory-level feedback which consists of non-accumulating and well-defined values.

\textbf{Therefore, we question whether variance reduction is necessary in optimizing the value model in RLHF.}

Based on the GAE computation in Equation \ref{eq:gae_computation}, we can rewrite the equation to obtain the value optimization target $V^{\text{target}}$:

\begin{equation}
    V^{\text{target}}(s_t) =
    \left\{
    \begin{array}{ll}
        \sum_{l = 0}^{T-t-1}\lambda^l (r_{t + l}+  V(s_{t + l + 1})-V(s_{t + l}))+V(s_t) ,& \lambda < 1.0 \\
        \sum_{l = 0}^{T-t-1} r_{t + l} ,& \lambda = 1.0
    \end{array}
    \right.
    \label{eq:gae_different_lambda}
\end{equation}

According to Equation \ref{eq:gae_different_lambda}, the reward assigned at the \texttt{<EOS>} token decays at a rate of $\lambda$ when propagating to the preceding tokens during GAE computation. The reward signal propagated to the $t$-th token is $\lambda^{T-t} r_{\text{<EOS>}}$. When $T-t$ is large, the resulting reward signal is essentially zero. With $\lambda = 1.0$, such reward signal decay would not occur, which makes it a desirable option for value optimization. Moreover, when $\lambda < 1.0$, value prediction is incorporated into the construction of the regression target. This approach belongs to semi-gradient descent methods, which tend to be unstable. Conversely, when $\lambda = 1.0$, the value is simply regressing to the accumulated rewards, resulting in a stable gradient-descent optimization.

\begin{figure}[ht]
    \centering
    \includegraphics[width=0.8\linewidth,height=0.4\textwidth]{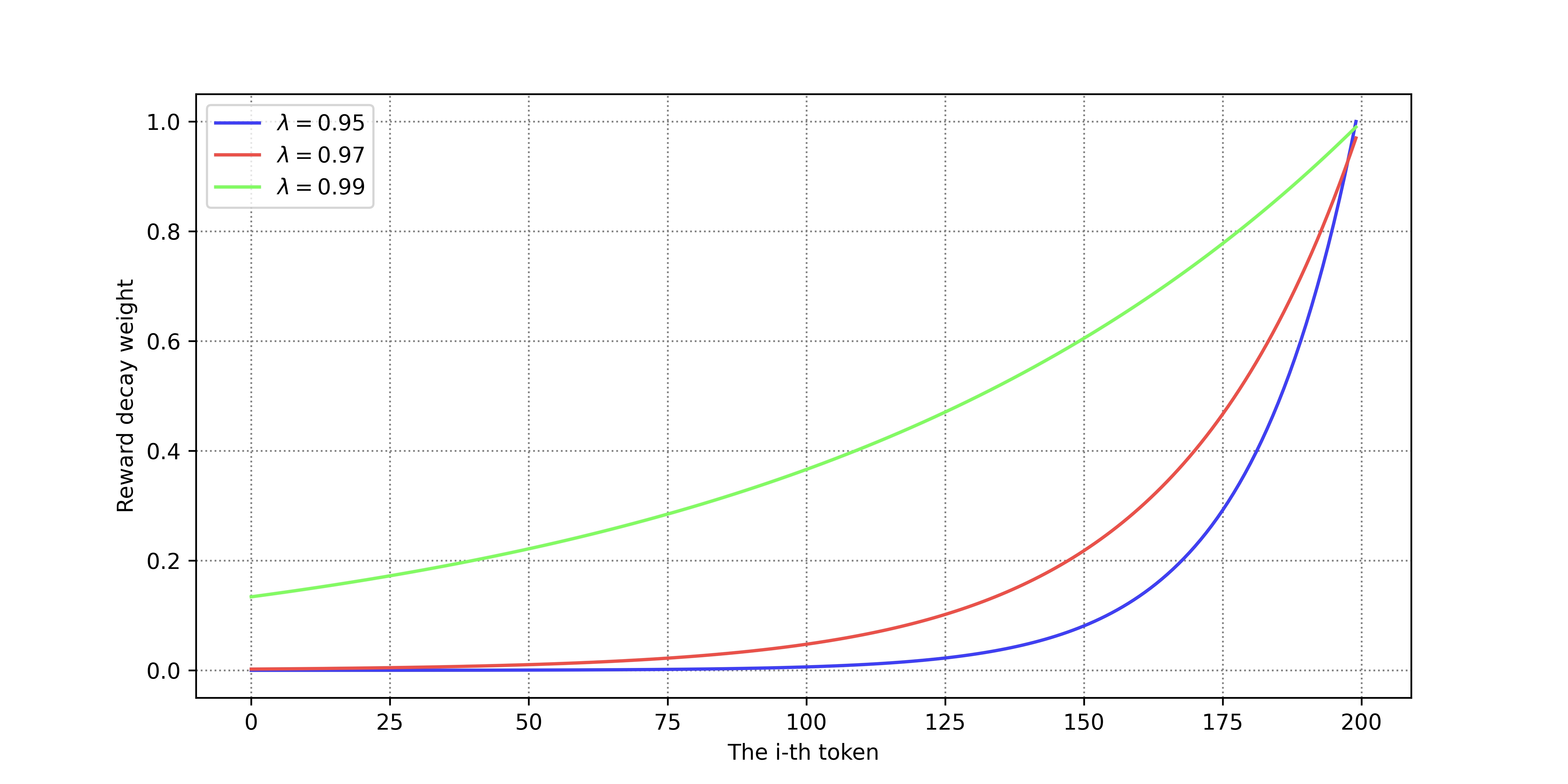}
    \caption{Reward signal decays as it propagates to preceding tokens.}
    \label{fig:reward_signal_decaying}
\end{figure}

In Figure \ref{fig:reward_signal_decaying}, we show that with $\lambda < 1.0$, the reward signal rapidly decays during propagation and preceding tokens are unable to receive the signal from the reward model. This phenomenon is exacerbated in tasks that require long CoT because the trajectory lengths are substantially longer. Therefore, optimizing the value in an unbiased manner outweighs learning it in a variance-reduced way because of the trajectory-level reward signal in RLHF. A similar argument is also proposed in \citep{ahmadian2024basicsrevisitingreinforcestyle}.

\textbf{However, variance reduction might still be necessary in policy optimization.}

Training large language models consumes a vast amount of computational resources. Under the constraints of time and computing power, achieving faster convergence during training is highly desirable. In PPO, the $\lambda$ parameter in GAE plays a crucial role in the bias-variance trade-off during policy updates. The variance of policy update can be analyzed in terms of the variances of the TD errors. Let \(\text{Var}[\delta_t]\) denote the variance of the TD error at time step \(t\). The variance of \(A_t^{\lambda}\) can be roughly computed as:

\begin{equation}
\begin{split}
\text{Var}[A_t^{\lambda}] &=\text{Var}\left[\sum_{l = 0}^{T-t-1}\lambda^l\delta_{t + l}\right] \\
& =\sum_{l = 0}^{T-t-1}\lambda^{2l}\text{Var}[\delta_{t + l}]+2\sum_{i = 0}^{T-t-1}\sum_{j = i + 1}^{T-t-1}\lambda^{i + j}\text{Cov}[\delta_{t + i},\delta_{t + j}],
\end{split}
\label{eq:gae_variance_analysis}
\end{equation}

where \(\text{Cov}[\delta_{t + i},\delta_{t + j}]\) is the covariance between the TD errors at time steps \(t + i\) and \(t + j\). Since \(\lambda\in[0, 1]\), as \(\lambda\) decreases, the weights \(\lambda^l\) for the later TD errors decrease more rapidly. This means that the contribution of the more variable and less reliable later TD errors to the overall advantage estimate is reduced, thereby reducing the variance of the advantage estimate.

Nevertheless, adjusting this $\lambda$ can have an additional impact on value optimization. To address this issue, we introduce \textbf{Decoupled-GAE}. This approach allows the policy to adopt a different $\lambda$ value from that of the value function. By doing so, the policy can better balance its own bias-variance trade-off, thereby enhancing the training efficiency.

Next, we show that using a value function obtained with a different $\lambda$ from the policy is mathematically justifiable without introducing additional bias. Let $\bar{V}$ represent the value estimate obtained with a potentially different $\lambda$, and define the $n$-step return with $\bar{V}$ as $G$:
\begin{equation}
    G_{t:t + h} =
    \left\{
    \begin{array}{ll}
        \sum_{l = 0}^{h-1} r_{t + l}+\bar{V}(s_{t + h}),& t + h < T \\
        \sum_{l = 0}^{T-h} r_{t + l},& t + h = T
    \end{array}
    \right.
    \label{eq:n-step_return_def}
\end{equation}

Then, the policy gradient with an arbitrary $\lambda$ can be rewritten as follows:

\begin{equation}
\begin{split}
\mathbb{E}_t \left[\nabla_{\theta}\log \pi_{\theta}(a_t|s_t) A_t \right] 
&= \mathbb{E}_t \left[\nabla_{\theta}\log \pi_{\theta}(a_t|s_t) \sum_{l = 0}^{T-t-1} \lambda^l (r_{t + l}+\bar{V}(s_{t + l + 1})-\bar{V}(s_{t + l})) \right] \\
&= \mathbb{E}_t \left[\nabla_{\theta}\log \pi_{\theta}(a_t|s_t) \left((1-\lambda)\sum_{l = 1}^{T-t-1} \lambda^{l-1}G_{t:t + l}+\lambda^{T-t-1}G_{t:T}-\bar{V}(s_t)\right) \right] \\
&= \mathbb{E}_t \left[\nabla_{\theta}\log \pi_{\theta}(a_t|s_t) \left((1-\lambda)\sum_{l = 1}^{T-t-1} \lambda^{l-1}G_{t:t + l}+\lambda^{T-t-1}G_{t:T}\right) \right] \\
\end{split}
\label{eq:gae_bias_derivation}
\end{equation}

Based on Equation \ref{eq:gae_bias_derivation}, plugging in an arbitrary value function does not introduce additional bias to the policy gradient. Given the substantial time and computational resources required for large language models, it is desirable to use a smaller $\lambda$ to expedite the convergence of the policy. A potential configuration could be $\lambda_{\text{policy}} = 0.95$ and $\lambda_{\text{value}} = 1.0$.

By combining Value-Pretraining and Decoupled-GAE, we propose Value-Calibrated Proximal Policy Optimization (VC-PPO) as shown in Alg \ref{alg:vc-ppo}, which is a simple yet effective approach to enhance PPO's performance in long CoT tasks. The main differences between VC-PPO and baseline PPO are highlighted.

\begin{algorithm}[H]
    \caption{Value-Calibrated Proximal Policy Optimization (VC-PPO)}
    \label{alg:vc-ppo}
    \begin{algorithmic}[1]
        \State \textbf{Input}: Initial policy $\pi_{\theta}$, \textcolor{red}{Pretrained value function $V_{\phi}$}, number of epochs $E$, number of mini-batches $M$, learning rate $\alpha_{\theta}$, learning rate $\alpha_{\phi}$, clip parameter $\epsilon$, \textcolor{red}{actor lambda $\lambda_\text{actor}$, critic lambda $\lambda_\text{critic}$}
        \State \textbf{Output}: Optimized policy $\pi_{\theta}$, optimized value function $V_{\phi}$
        \For{$e = 1$ \textbf{to} $E$}
            \State Collect a set of trajectories $\tau=\{(s_t, a_t, r_t)\}_{t = 0}^{T-1}$ using the current policy $\pi_{\theta}$
            \State \textcolor{red}{Compute the advantages $A$ with $\lambda_\text{actor}$}
            \State \textcolor{red}{Compute the value targets $R$ with $\lambda_\text{critic}$}
            \State Split the collected data into $M$ mini-batches
            \For{$m = 1$ \textbf{to} $M$}
                \State Sample a mini-batch of data $(s, a, A, R)$ from the collected data
                \State Compute the probability ratio $r(\theta)=\frac{\pi_{\theta}(a|s)}{\pi_{\theta_{\text{old}}}(a|s)}$
                \State Compute the clipped surrogate objective $\mathcal{L}^{CLIP}(\theta)=\min\left(r(\theta)A, \text{clip}(r(\theta), 1-\epsilon, 1+\epsilon)A\right)$
                \State Compute the value function loss $\mathcal{L}^{VF}(\phi)=\frac{1}{2}(V_{\phi}(s)-R)^2$
                \State Maximizing $\mathcal{L}^{CLIP}(\theta)$ using: $\theta\leftarrow\theta+\alpha_{\theta}\nabla_{\theta}\mathcal{L}^{CLIP}(\theta)$
                \State Minimizing $\mathcal{L}^{VF}(\phi)$ using: $\phi\leftarrow\phi-\alpha_{\phi}\nabla_{\phi}\mathcal{L}^{VF}(\phi)$
            \EndFor
        \EndFor
        \State \Return $\pi_{\theta}$, $V_{\phi}$
    \end{algorithmic}
\end{algorithm}
\section{Experiments}

\subsection{Setup}
\textbf{Datasets.} To comprehensively demonstrate the effectiveness of our proposed algorithm, we conduct experiments on American Invitational Mathematics Examination (AIME) problems, which typically demand a long chain of thought for solution. The test set consists of AIME questions from the most recent two years. Conversely, the training set is composed of questions from all past AIME competitions, supplemented with some artificially constructed difficult mathematical problems. To evaluate the model's generalizability, we simultaneously monitor its performance in typical long CoT scenarios, such as General-Purpose Question-Answering (GPQA) \cite{gpqa} and Codeforces.

\textbf{Cold Start.} This phase aims to enhance the model's reasoning capabilities within a specific reasoning format. We used dozens of samples with a format that requires the model to place its thinking process between \texttt{<think>} and \texttt{</think>} tags before presenting the final answer. These samples were used to fine-tune the Qwen2.5 32B model \citep{qwen2.5}, which we employ in our experiments for better reproducibility.

\textbf{Reward Modeling.} We adopt the methodology commonly used in classical reasoning tasks across domains such as mathematics, code, and logical reasoning. This approach utilizes rule-based rewards to guide the learning process. When assigning the reward score, the verifier ignores the thinking part enclosed by the \texttt{<think>} tokens and extracts only the answer part for evaluation. Correct answers are assigned a score of $1.0$, while incorrect answers receive a score of $- 1.0$.

\textbf{RL Baseline.} 
In our experiments, we use verl \citep{sheng2024hybridflow} as our experimental framework. The Proximal Policy Optimization (PPO) algorithm described in \citep{ouyang2022training} serves as the baseline, with $\lambda$ set to 0.95 by default. The learning rates for the policy model and the value model are set to $1\times10^{-6}$ and $2\times10^{-6}$, respectively. The KL penalty coefficient is set to 0 because the rule-based reward cannot be hacked in the same way as that of a general reward model. We adopt different context length settings of 8k and 16k for different purposes: the 16k setting is used for comparison with state-of-the-art results, and the 8k setting is used for ablation studies.

\textbf{Value-Pretraining.} We freeze the policy model and set the Generalized Advantage Estimation (GAE) $\lambda$ to 1.0 to obtain an unbiased return. The other hyperparameters are the same as those of the baseline Proximal Policy Optimization (PPO). By saving the value model at different steps of value pretraining, we can acquire multiple initial value checkpoints for RL training. We also conduct ablation studies on these checkpoints in our experiments.

\textbf{Decoupled-GAE.} Due to the value oscillation and reward signal decay described in Section \ref{subsec:decoupled-gae}, $\lambda_{\text{critic}}$ is set to 1.0. Meanwhile, $\lambda_{\text{actor}}$ used in the policy is maintained at 0.95 to enable a fair comparison with the baseline PPO. Subsequently, we assign values to $\lambda_{\text{actor}}$ ranging from 0.9 to 1.0 to investigate its impact on the convergence of the policy, while $\lambda_{\text{critic}}$ remains at 1.0.


\subsection{Experimental Results}

We conduct RL training on the Qwen-32B-Base model using the proposed Value-Calibrated Proximal Policy Optimization (VC-PPO) algorithm. We then compare our model with the well-established Generalized Proximal Policy Optimization (GRPO) algorithm, which is employed in the DeepSeek-R1 model \citep{deepseekai2025deepseekr1incentivizingreasoningcapability}. This experiment utilizes a 16k context length to attain the state-of-the-art performance.

The results are presented in Table \ref{tab:release_exp}. The proposed VC-PPO algorithm significantly outperforms GRPO under the same experimental setting. Our primary objective is to optimize the model's performance on the American Invitational Mathematics Examination (AIME), which consists of Olympiad-level math problems. Consequently, the majority of the training data is math-related, and VC-PPO demonstrates the most substantial advantage on the AIME dataset.

\begin{table}[H]
\centering
\begin{tabular}{c|cccc}
    \toprule
    & \multicolumn{2}{c}{AIME 2024} & GPQA & CodeForces \\
    Model & pass@1 & pass@32 & pass@1 & pass@1  \\
    \midrule
    GRPO & 38.9 & 70.0 & \textbf{49.4} & 12.6 \\
    VC-PPO & \textbf{48.8} & \textbf{73.3} & 48.8 & \textbf{12.8}  \\
    \bottomrule
\end{tabular}
\caption{Comparison between VC-PPO and GRPO in 16K context length.}
\label{tab:release_exp}
\end{table}

\begin{figure}[ht]
    \centering
    \includegraphics[width=0.8\linewidth]{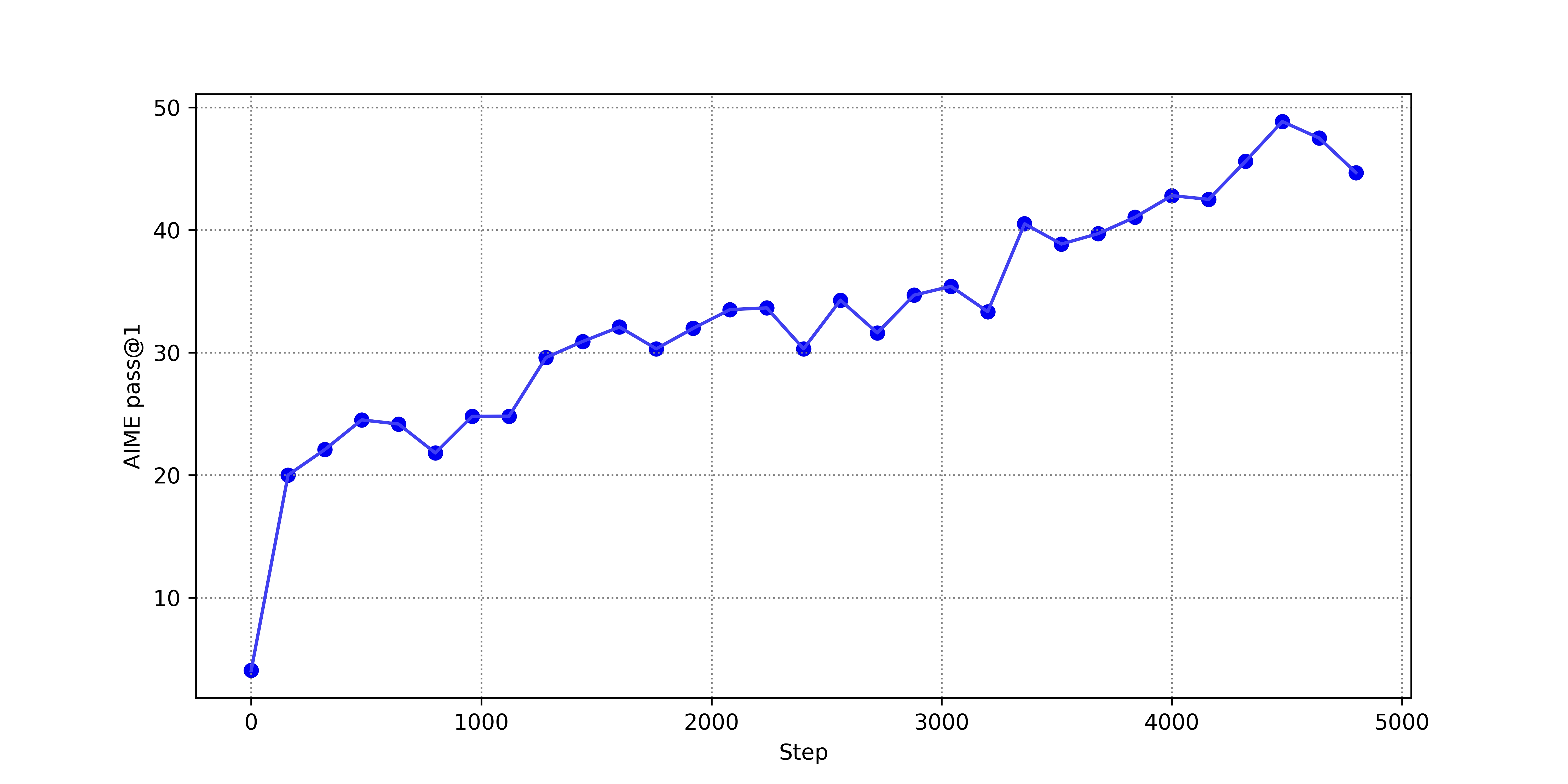}
    \caption{AIME accuracy during training.}
    \label{fig:AIME_vcppo}
\end{figure}

To the best of our knowledge, a pass@1 score of 48.8 on the AIME dataset stands as the highest performance attained by a Qwen-32B-Base model without employing distillation techniques. This score surpasses the AIME score of 47.0 reported in the DeepSeek-R1 technical report \citep{deepseekai2025deepseekr1incentivizingreasoningcapability} under comparable experimental settings\footnote{It should be noted that a direct comparison between these two results is not entirely feasible because the dataset used for RL training in DeepSeek-R1 has not been made publicly available.}. The increasing pass rate of the AIME dataset during the training process is illustrated in Figure \ref{fig:AIME_vcppo}. Additionally, we have deployed the VC-PPO algorithm in our internal model, which has achieved an AIME score of 74.

\textbf{For ablation studies, we use an 8k context length to enhance training efficiency.}

In Table \ref{tab:main_exp}, we showcase the ablation results of Value-Pretraining and Decoupled-GAE. When directly applying the Proximal Policy Optimization (PPO) algorithm, it fails to improve the performance of the pre-trained model. This is because the output length of the model collapses. In contrast, the proposed Value-Calibrated Proximal Policy Optimization (VC-PPO) algorithm demonstrates a significant performance boost, highlighting its superiority in handling tasks that demand a long CoT.

Moreover, when we conduct ablation experiments by removing either the Value-Pretraining or the Decoupled-GAE component from VC-PPO, there is a notable drop in performance. This decline emphasizes the crucial roles that both Value-Pretraining and Decoupled-GAE play in the effectiveness of our proposed VC-PPO algorithm.

\begin{table}[H]
\centering
\begin{tabular}{c|cccc}
    \toprule
    & \multicolumn{2}{c}{AIME} & GPQA & CodeForces\\
    Alg. & pass@1 & pass@32 & pass@1 & pass@1  \\
    \midrule
    GRPO & 35.8 & 63.0 &-&-\\
    PPO & 5.6 & 36.7 & 38.7 & 7.3 \\
    VC-PPO w/o Decoupled-GAE & 29.4 & 66.7 & 46.9 & 9.9  \\
    VC-PPO & \textbf{41.9} & \textbf{76.6} & \textbf{48.6} & \textbf{11.4}  \\
    \bottomrule
\end{tabular}
\caption{Ablation study on VC-PPO's components.}
\label{tab:main_exp}
\end{table}

In Figure \ref{tab:pretrain_results}, we compare the model performance at the same step in this ablation experiment, specifically after 100 steps of training. The decision to conduct this comparison is driven by the evident divergence in performance trends. The optimal configuration involves pretraining the value model for 100 steps. This is because additional training beyond this point might induce overfitting, which could negatively impact the model's generalization ability.

\begin{table}[H]
\centering
\begin{tabular}{c|cccc}
    \toprule
    & \multicolumn{2}{c}{AIME} & GPQA & CodeForces \\
    Value Pretraining Steps. & pass@1 & pass@32 & pass@1 & pass@1  \\
    \midrule
    50 & 20.6 & 56.7 & 43.9 & 7.7 \\
    100 & \textbf{30.1} & 63.3 & \textbf{48.5} & 8.6 \\
    150 & 25.1 & \textbf{63.6} & 48.4 & \textbf{8.9} \\
    \bottomrule
\end{tabular}
\caption{Ablation study on Value-Pretraining steps.}
\label{tab:pretrain_results}
\end{table}

The results of the ablation study on $\lambda_{\text{actor}}$ are presented in Table \ref{tab:lambda_results}. It should be highlighted that all experimental groups with $\lambda_{\text{actor}} < 1.0$ significantly outperform the group with $\lambda_{\text{actor}} = 1.0$. This outcome supports our analysis in Section \ref{subsec:decoupled-gae}. In the case of the American Invitational Mathematics Examination (AIME), the experimental group with $\lambda_{\text{actor}} = 0.99$ outperforms the other groups with lower $\lambda_{\text{actor}}$ values. However, there is only a slight decrease in performance within a certain range between $0.95$ and $1.0$. Therefore, the recommended setting for $\lambda_{\text{actor}}$ is $\lambda_{\text{actor}} \in [0.95, 1.0)$.

\begin{table}[H]
\centering
\begin{tabular}{c|cccc}
    \toprule
    & \multicolumn{2}{c}{AIME} & GPQA & CodeForces \\
    $\lambda_{\text{actor}}$ & pass@1 & pass@32 & pass@1 & pass@1  \\
    \midrule
    0.9 & 34.3 & 70.0 & 48.1 & 11.7 \\
    0.95 & 41.3 & 73.3 & \textbf{48.7} & \textbf{12.8} \\
    0.99 & \textbf{41.9} & \textbf{76.7} & 48.3 & 11.4  \\
    1.0 & 29.4 & 66.7 & 46.9 & 8.4 \\
    \bottomrule
\end{tabular}
\caption{Ablation study on different $\lambda_{\text{actor}}$ values.}
\label{tab:lambda_results}
\end{table}

\subsection{Discussion}

\textbf{A smooth initial state for training is crucial in RLHF, especially in long CoT tasks.}

In traditional RL, both the value function and the policy are typically initialized randomly. However, in RLHF, the initial policy is usually initialized from the supervised fine-tuning (SFT) policy. This SFT policy acts as a strong prior for the learning process. In long CoT tasks, the initial policy is further enhanced with the CoT pattern, offering an even stronger prior.

\begin{figure}[htbp]
    \centering
    \begin{subfigure}[b]{0.45\textwidth}
        \centering
        \begin{center}
            \includegraphics[width=1.0\linewidth]{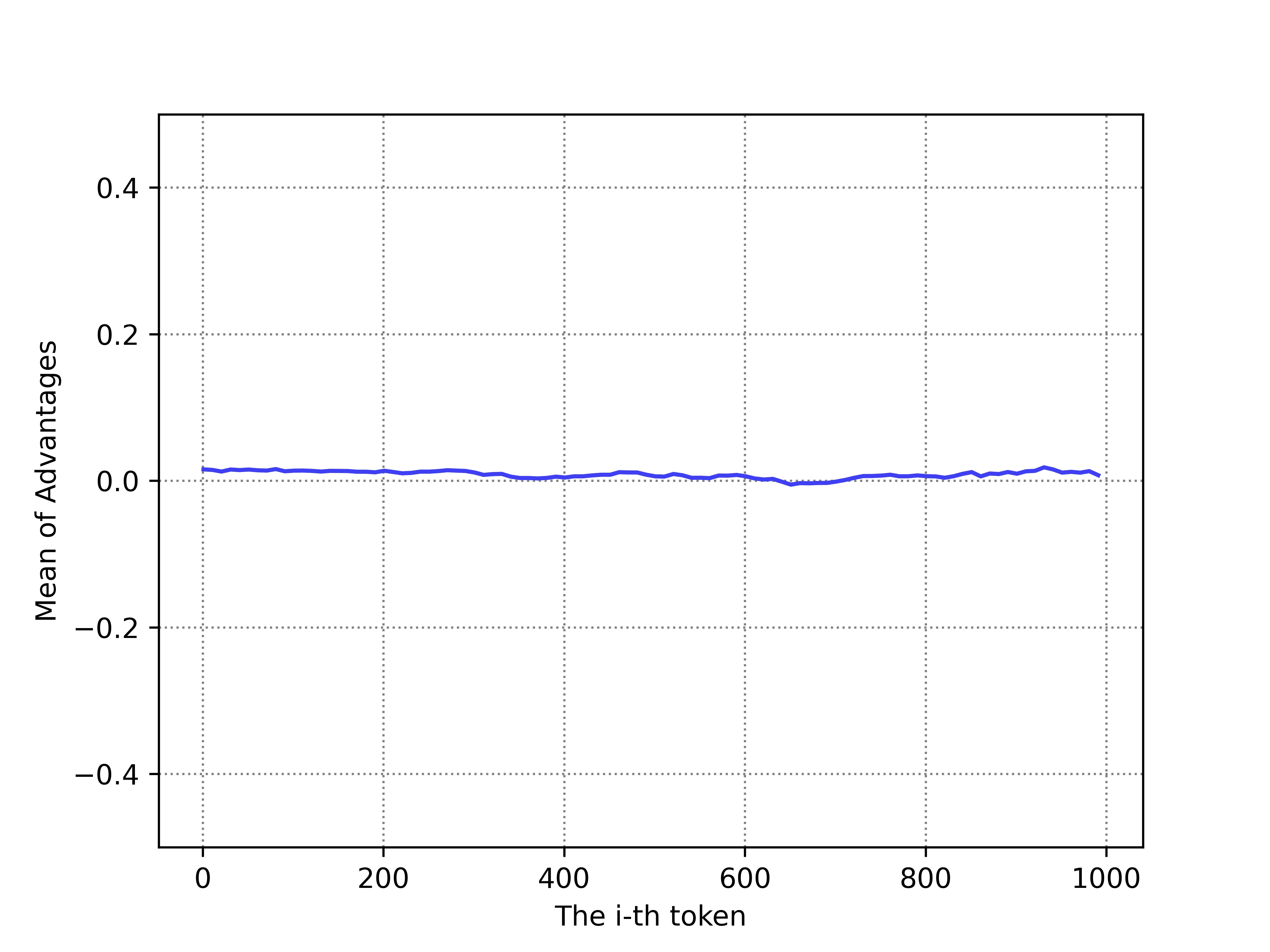}
        \end{center}
        \caption{Advantages at different token positions}
        \label{fig:ppo_failure_fixed_2}
    \end{subfigure}
    \hfill
    \begin{subfigure}[b]{0.45\textwidth}
        \centering
        \includegraphics[width=1.0\linewidth]{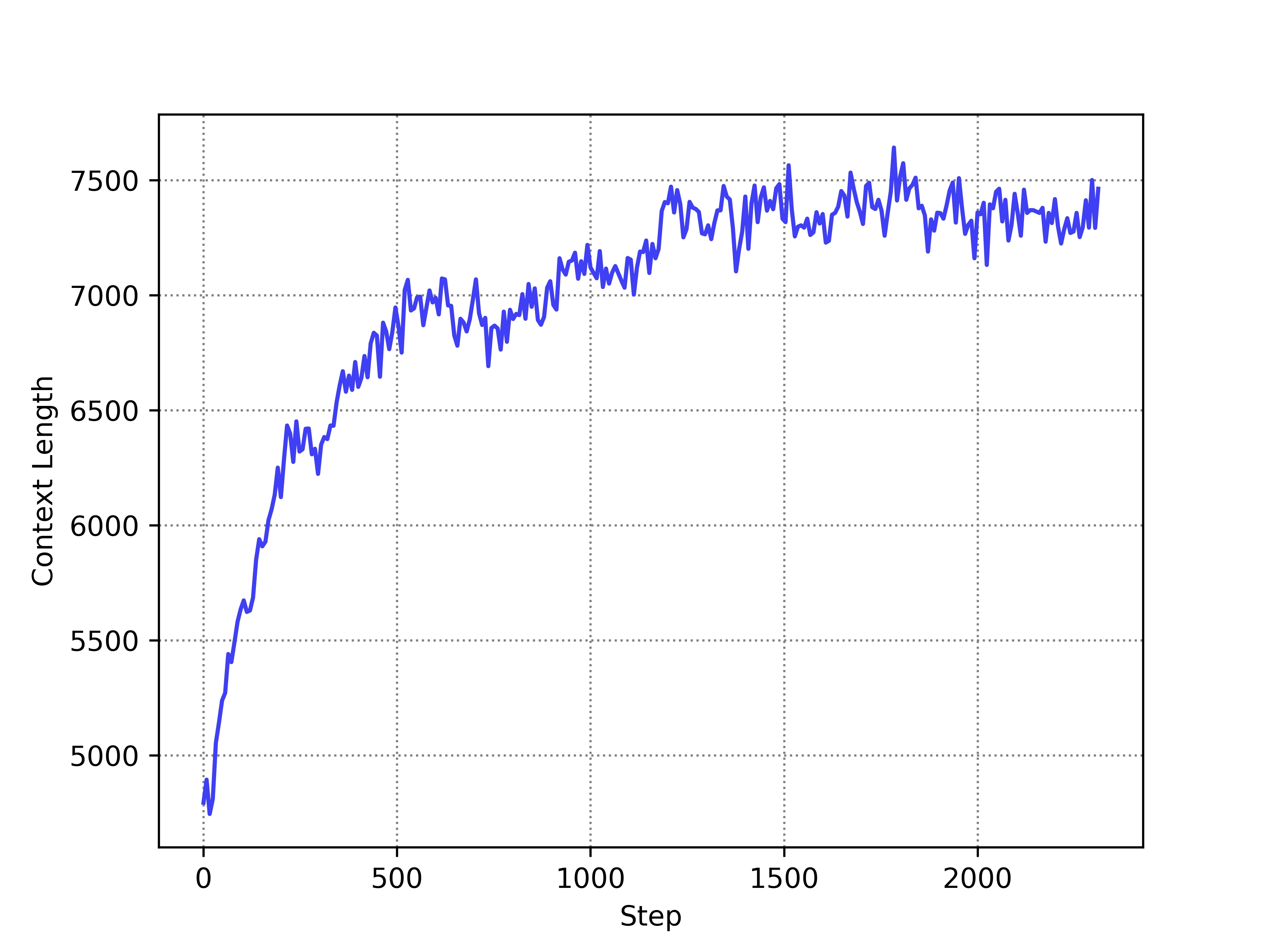}
        \caption{Model output length}
        \label{fig:ppo_failure_fixed_3}
    \end{subfigure}
    \caption{Advantage estimate and output length after value-pretraining.}
    \label{fig:ppo_failure_fixed}
\end{figure}

Our empirical observations suggest that as the prior policy becomes stronger, it is increasingly essential to align the value model with the policy. Otherwise, the painstakingly constructed CoT pattern can easily be disrupted, as demonstrated in Figure \ref{fig:ppo_failure_modes}. In our experiment, after applying the value-pretraining technique, which effectively aligns the value model with the initial policy, the collapse in output length is no longer observed. This result clearly highlights the significance of having a fully-aligned value model, as shown in Figure \ref{fig:ppo_failure_fixed}.

\textbf{Value-Pretraining injects knowledge into the value model, which is a superior form of value warm-up.}

\begin{figure}[ht]
    \centering
    \includegraphics[width=0.45\linewidth]{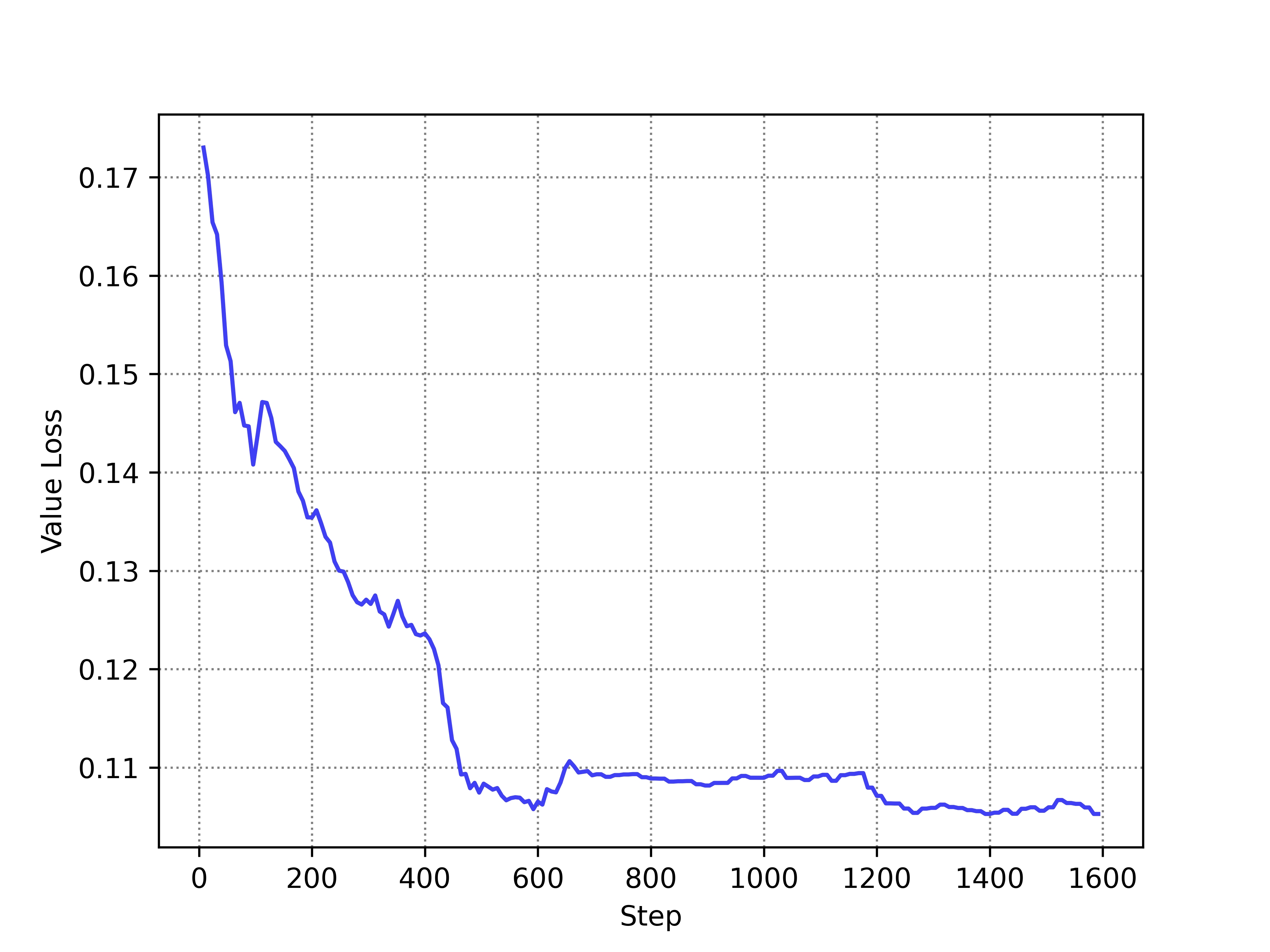}
    \caption{Value-pretraining loss.}
    \label{fig:pretraining_value_loss}
\end{figure}

We present the value loss during value-pretraining in Figure \ref{fig:pretraining_value_loss}, where a two-stage convergence pattern can be observed. In the first stage, there is a rapid decline in value loss. We interpret this stage as range alignment, which shares similarities with the commonly used value warm-up technique in RL. However, in the second stage, the value loss decreases at a slower pace. We interpret this stage as knowledge injection. In this stage, the model starts to learn which tokens are more advantageous, a crucial aspect that has often been overlooked in previous research. As shown in Table \ref{tab:pretrain_results}, this stage has a substantial impact on the final performance of our model.

\textbf{Value optimization dynamics are more tolerant to variance, which leads to different variance-bias preferences in value and policy.}

Based on the experimental results presented in Table \ref{tab:main_exp} and the ablation results in Table \ref{tab:lambda_results}, we can conclude that the value model favors a larger $\lambda$, resulting in higher variance but lower bias. In contrast, the policy model prefers lower variance which means a lower $\lambda$ than $1.0$. Notably, a relatively lower bias is still required at the same time since the extremely variance can hurt the performance anyway. This finding implicitly suggests that regression-style loss objectives, such as the mean squared error (MSE) loss used in value optimization, are less sensitive to variance. Conversely, policy-gradient-style objectives are more likely to be adversely affected by variance. This could serve as a promising avenue for further research in RL or RLHF.

\section{Conclusion}
In this study, we delved into the failure of PPO in long CoT tasks and proposed VC-PPO as a solution. By identifying value initialization bias and reward signal decay as the main problems, we introduced value pretraining and decoupled-GAE techniques. Value pretraining aligns the value model with the initial policy, preventing the loss of the CoT pattern and improving performance. Decoupling the GAE computation for the policy and value allows for better bias-variance trade-offs in both components.
Experimental results on AIME, CodeForces, and GPQA datasets show that VC-PPO outperforms the baseline PPO significantly. Ablation studies further emphasize the crucial role of value pretraining and decoupled-GAE in VC-PPO. Additionally, our research reveals differences in variance-bias preferences between value and policy models, which could be a promising area for future RL and RLHF research. Overall, VC-PPO provides an effective way to enhance PPO's performance in long CoT tasks, contributing to the advancement of LLMs in complex reasoning tasks.

\clearpage

\bibliographystyle{plainnat}
\bibliography{main}



\end{document}